\documentclass[conference]{IEEEtran}
\usepackage{cite}
\usepackage{amsmath,amssymb,amsfonts}
\usepackage{graphicx}
\usepackage{textcomp}
\usepackage{xcolor}
\usepackage{listings}
\usepackage{booktabs}
\usepackage{multirow}
\usepackage{url}
\usepackage{balance}
\usepackage{enumitem}
\usepackage{tikz}
\usetikzlibrary{shapes,arrows.meta,positioning,calc,fit}
\usepackage{amsthm}
\newtheorem{definition}{Definition}

\lstdefinelanguage{Rust}{
  keywords={fn, let, mut, struct, impl, trait, pub, use, self, Self, for, in, if, else, match, return, async, await, where, type, mod, crate, super, dyn, Box, Result, Ok, Err, Some, None, true, false},
  keywordstyle=\color{blue}\bfseries,
  ndkeywords={Aspect, JoinPoint, ProceedingJoinPoint, AspectError, String, u64, bool, Duration, Vec, HashMap, Arc, Mutex, HashSet},
  ndkeywordstyle=\color{purple},
  sensitive=true,
  comment=[l]{//},
  morecomment=[s]{/*}{*/},
  commentstyle=\color{gray}\itshape,
  stringstyle=\color{red},
  morestring=[b]",
  basicstyle=\ttfamily\scriptsize,
  breaklines=true,
  frame=single,
  numbers=left,
  numberstyle=\tiny\color{gray},
  tabsize=2,
  xleftmargin=0.2cm
}
\newcommand{\istar}{$i^*$}
\newcommand{\aspectrs}{\textsc{aspect-rs}}
\newcommand{\agentfw}{\textsc{ZeroClaw}}
\newcommand{\patname}[1]{\textsc{#1}}

\begin{document}

\title{From Goals to Aspects, Revisited:\\An NFR Pattern Language for Agentic AI Systems}

\author{\IEEEauthorblockN{Yijun Yu}
\IEEEauthorblockA{The Open University\\
Milton Keynes, UK\\
yijun.yu@open.ac.uk}}

\maketitle

\begin{abstract}
Agentic AI systems exhibit numerous crosscutting concerns---security, observability, cost management, fault tolerance---that are poorly modularized in current implementations, contributing to the high failure rate of AI projects in reaching production. The goals-to-aspects methodology proposed at RE 2004 demonstrated that aspects can be systematically discovered from \istar{} goal models by identifying non-functional softgoals that crosscut functional goals. This paper revisits and extends that methodology to the agentic AI domain. We present a pattern language of 12 reusable patterns organized across four NFR categories (security, reliability, observability, cost management), each mapping an \istar{} goal model to a concrete aspect implementation using an AOP framework for Rust. Four patterns address agent-specific crosscutting concerns absent from traditional AOP literature: tool-scope sandboxing, prompt injection detection, token budget management, and action audit trails. We extend the V-graph model to capture how agent tasks simultaneously contribute to functional goals and non-functional softgoals. We validate the pattern language through a case study analyzing an open-source autonomous agent framework, demonstrating how goal-driven aspect discovery systematically identifies and modularizes crosscutting concerns. The pattern language offers a principled approach for engineering reliable agentic AI systems through early identification of crosscutting concerns.
\end{abstract}


\section{Introduction}
\label{sec:intro}

Agentic AI systems---autonomous software agents powered by large language models (LLMs) that can plan, reason, and execute tool-calling workflows---represent one of the most rapidly growing areas of software engineering~\cite{wang2024survey}. These systems go beyond simple chatbots: they make autonomous decisions, invoke external tools (shell commands, file operations, web searches), and manage multi-turn conversations with minimal human oversight. Industry reports indicate that inquiries about multi-agent systems have surged dramatically, yet the vast majority of AI projects fail to reach production~\cite{venturebeat2019ai}.

A critical but underexplored contributor to this failure is the \emph{inadequate engineering of non-functional requirements (NFRs)}. Agentic AI systems exhibit numerous crosscutting concerns---security enforcement, observability, cost management, fault tolerance---that pervade every module of the system. In current agent frameworks, these concerns are implemented in an ad-hoc manner: authorization checks are scattered across tool handlers, logging is inconsistently applied, rate limiting is duplicated, and cost controls are often absent entirely. This \emph{scattering} and \emph{tangling} of crosscutting concerns is precisely the problem that aspect-oriented software development (AOSD) was designed to solve~\cite{kiczales2001overview}.

Twenty-two years ago, Yu et al.\ proposed a systematic methodology for discovering aspects from \istar{} goal models at RE 2004~\cite{yu2004goals}. That work demonstrated that non-functional softgoals in goal models naturally crosscut functional goals, and that these crosscutting relationships can be systematically identified and extracted as \emph{aspects}. The methodology was illustrated through a media shop case study, identifying security, usability, and portability as crosscutting aspects. Subsequent work extended this to the full software lifecycle~\cite{niu2009aspects} and developed tool support through EA-Miner~\cite{sampaio2005eaminer}.

In this paper, we revisit and extend the goals-to-aspects methodology to the emerging domain of agentic AI. We argue that agent systems present both a compelling new application domain for aspect-oriented requirements engineering (AORE) and introduce \emph{novel} crosscutting concerns---such as prompt injection defense, token budget management, and tool-scope sandboxing---that are absent from the traditional AOP literature. Moreover, the \emph{intensity} of crosscutting in agent systems exceeds that of traditional software: a single agent task like ``Call LLM Provider'' simultaneously touches security, cost, reliability, and observability concerns, creating a density of overlapping crosscutting that the original V-graph model can capture but that existing case studies have not addressed.

Our contributions are:
\begin{enumerate}[leftmargin=*]
  \item A \textbf{pattern language} of 12 reusable patterns organized across four NFR categories, each mapping an \istar{} goal model to a concrete aspect implementation in Rust. Four patterns address agent-specific crosscutting concerns not found in traditional AOP.
  \item An \textbf{extension of the V-graph model}~\cite{yu2004goals} to the agentic AI domain, capturing how agent tasks simultaneously contribute to functional goals (tool execution, LLM interaction) and non-functional softgoals (safety, cost efficiency, compliance).
  \item A \textbf{case study} analyzing an open-source autonomous agent framework (192 files, 129{,}040~LOC), quantifying the degree of concern scattering across eleven crosscutting concerns (seven established, four newly identified) and demonstrating that V-graph analysis identifies both scattered implementations and missing NFR coverage.
\end{enumerate}

The remainder of this paper is organized as follows. Section~\ref{sec:background} reviews background on goal-oriented RE, AOSD, and agentic AI. Section~\ref{sec:methodology} presents our three-phase methodology. Section~\ref{sec:patterns} describes the pattern language. Section~\ref{sec:casestudy} presents the case study. Section~\ref{sec:discussion} discusses implications and threats to validity. Section~\ref{sec:related} surveys related work. Section~\ref{sec:conclusion} concludes.

\section{Background}
\label{sec:background}

\subsection{From Goals to Aspects (RE 2004)}
\label{sec:bg-goals}

Yu et al.~\cite{yu2004goals} proposed a methodology for discovering aspects from requirements goal models. The key insight is that non-functional requirements, modeled as \emph{softgoals} in the \istar{} framework~\cite{yu1997modelling}, naturally crosscut functional goals. When a softgoal such as ``Security'' contributes to (or is operationalized by) tasks that also serve multiple functional goals, this reveals a crosscutting concern that should be modularized as an aspect.

The methodology introduces the \emph{V-graph}, a characteristic structure in the goal model where:
\begin{itemize}[leftmargin=*]
  \item The left vertex represents a \emph{functional goal} (e.g., ``Process Order'');
  \item The right vertex represents an \emph{NFR softgoal} (e.g., ``Security'');
  \item The bottom vertex represents \emph{implementing tasks} that contribute to both.
\end{itemize}
The V-shape captures the essence of crosscutting: a single task serves two distinct concerns, one functional and one non-functional. The \emph{AspectFinder} algorithm systematically identifies such V-graphs in the goal model and extracts the corresponding aspects~\cite{yu2004goals}.

\subsection{The \istar{} Framework}
\label{sec:bg-istar}

The \istar{} (``i-star'') framework~\cite{yu1997modelling,dalpiaz2016istar} is a goal-oriented modeling language for early-phase requirements engineering. It models systems as networks of \emph{actors} with intentional properties---goals, softgoals, tasks, and resources---connected by \emph{dependency} relationships. Two complementary models are used:
\begin{itemize}[leftmargin=*]
  \item \textbf{Strategic Dependency (SD)}: Models dependencies between actors.
  \item \textbf{Strategic Rationale (SR)}: Models internal goal decomposition within actors through AND/OR refinement, means-ends links, and contribution links (positive~++, positive~+, negative~--, negative~-) to softgoals.
\end{itemize}

\subsection{Aspect-Oriented Programming}
\label{sec:bg-aop}

Aspect-oriented programming (AOP)~\cite{kiczales1997aop,kiczales2001overview} modularizes crosscutting concerns into \emph{aspects}. An aspect encapsulates behavior that cuts across multiple modules, applied at designated \emph{join points} through \emph{advice} (before, after, around). While AspectJ~\cite{kiczales2001overview} pioneered AOP for Java, the Rust ecosystem has lacked comprehensive AOP support until recently.

We leverage \aspectrs{}~\cite{aspectrs2025}, an AOP framework for Rust that provides: (1)~an \texttt{Aspect} trait with \texttt{before}, \texttt{after}, \texttt{after\_error}, and \texttt{around} advice types; (2)~compile-time weaving via procedural macros (\texttt{\#[aspect(...)]}); (3)~eight production-ready aspects for common crosscutting concerns; and (4)~near-zero runtime overhead through static dispatch. The framework's key abstraction is:

\begin{lstlisting}
pub trait Aspect: Send + Sync {
  fn before(&self, ctx: &JoinPoint) {}
  fn after(&self, ctx: &JoinPoint,
           result: &dyn Any) {}
  fn after_error(&self, ctx: &JoinPoint,
                 error: &AspectError) {}
  fn around(&self, pjp: ProceedingJoinPoint)
    -> Result<Box<dyn Any>, AspectError> {
    pjp.proceed()
  }
}
\end{lstlisting}

\subsection{Agentic AI Systems}
\label{sec:bg-agents}

An agentic AI system is an autonomous software agent that uses an LLM as its reasoning engine, augmented with tool-calling capabilities, memory, and multi-turn conversation management~\cite{wang2024survey}. Unlike traditional software, agents make \emph{autonomous decisions} about which tools to invoke, what information to retrieve, and how to synthesize results---often with minimal human oversight.

Common architectural patterns include ReAct (Reasoning + Acting)~\cite{yao2023react}, Reflexion~\cite{shinn2023reflexion}, function calling, and multi-agent orchestration~\cite{xi2024agentgym}. Agent frameworks typically provide: LLM provider integration (OpenAI, Anthropic, etc.), tool registries (shell, file I/O, web search), memory systems (conversation history, vector stores), and communication channels.

The crosscutting concerns in agent systems include but extend beyond traditional software: in addition to logging, authorization, and caching, agents require \emph{prompt injection defense} (OWASP \#1 for LLM applications~\cite{owasp2025llm}), \emph{token budget management} (controlling LLM API costs), \emph{tool-scope sandboxing} (containing agent actions within safe boundaries), and \emph{action audit trails} (recording agent decisions for accountability).

What distinguishes agent systems from traditional software is the \emph{density} of crosscutting: a single function call to an LLM provider simultaneously involves security (prompt injection risk), cost (token consumption), reliability (provider availability), observability (decision logging), and safety (output validation). This creates a unique engineering challenge where multiple NFRs must be enforced at the same join points, making aspect-oriented modularization particularly valuable.

\section{Methodology: Goals to Aspects for Agents}
\label{sec:methodology}

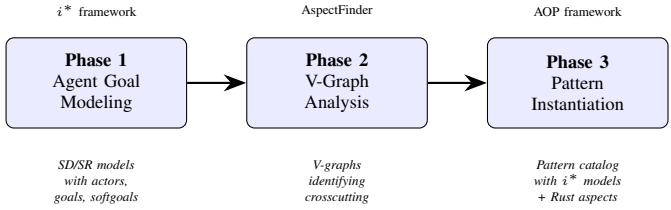
\begin{figure}[t]
\centering
\begin{tikzpicture}[
  phase/.style={rectangle, draw, rounded corners=3pt, minimum width=2.4cm, minimum height=1.2cm, font=\scriptsize, align=center, fill=blue!8},
  arrow/.style={-{Stealth[length=3mm]}, thick},
  label/.style={font=\tiny\itshape, text width=2.2cm, align=center}
]
\node[phase] (p1) at (0,0) {\textbf{Phase 1}\\Agent Goal\\Modeling};
\node[phase] (p2) at (3.2,0) {\textbf{Phase 2}\\V-Graph\\Analysis};
\node[phase] (p3) at (6.4,0) {\textbf{Phase 3}\\Pattern\\Instantiation};
\draw[arrow] (p1) -- (p2);
\draw[arrow] (p2) -- (p3);
\node[label, below=0.3cm of p1] {SD/SR models\\with actors,\\goals, softgoals};
\node[label, below=0.3cm of p2] {V-graphs\\identifying\\crosscutting};
\node[label, below=0.3cm of p3] {Pattern catalog\\with \istar{} models\\+ Rust aspects};
\node[font=\tiny, above=0.15cm of p1] {\istar{} framework};
\node[font=\tiny, above=0.15cm of p2] {AspectFinder};
\node[font=\tiny, above=0.15cm of p3] {AOP framework};
\end{tikzpicture}
\caption{Three-phase methodology for discovering aspects from agent goal models.}
\label{fig:methodology}
\end{figure}

We extend the RE 2004 goals-to-aspects methodology~\cite{yu2004goals} to the agentic AI domain through a three-phase process (Fig.~\ref{fig:methodology}).

\subsection{Phase 1: Agent Goal Modeling}
\label{sec:phase1}

We construct an \istar{} Strategic Dependency (SD) model identifying the key actors in an agentic AI system and their dependencies:

\begin{itemize}[leftmargin=*]
  \item \textbf{Agent User}: Depends on the agent for task execution, information retrieval, and communication.
  \item \textbf{Agent System}: The autonomous agent, depending on LLM providers for reasoning and tool providers for action execution.
  \item \textbf{LLM Provider}: External service providing text generation and reasoning capabilities.
  \item \textbf{Tool Provider}: Services providing shell execution, file I/O, web search, and other capabilities.
  \item \textbf{Operator}: Depends on the agent system for security, monitoring, cost control, and compliance.
\end{itemize}

For each actor, we construct a Strategic Rationale (SR) model decomposing goals into subgoals and tasks through AND/OR refinement. Crucially, we identify both \emph{functional goals} (what the agent does) and \emph{NFR softgoals} (how well it does it), and trace contribution links between tasks and softgoals.

The agent domain introduces characteristic goal decompositions not present in traditional software. For example, the functional goal \emph{Execute Tool} AND-decomposes into \emph{Parse Tool Call}, \emph{Validate Parameters}, \emph{Execute Action}, and \emph{Return Result}. Each subtask potentially contributes to multiple NFR softgoals: \emph{Validate Parameters} contributes positively ($++$) to Security and Safety, while \emph{Execute Action} contributes negatively ($--$) to Security (risk of unauthorized action) and Cost (resource consumption). These multi-directional contribution links generate the overlapping V-graphs described in Phase~2.

\subsection{Phase 2: V-Graph Analysis for Agents}
\label{sec:phase2}

We extend the V-graph model~\cite{yu2004goals} to define a \textbf{V-graph for Agents}:

\begin{definition}[V-graph for Agents]
A V-graph for agents $V = (g_f, g_{nf}, T)$ is a subgraph of the \istar{} SR model consisting of:
\begin{itemize}[leftmargin=*]
  \item A \emph{functional goal} $g_f$ (e.g., ``Execute Tool'');
  \item An \emph{NFR softgoal} $g_{nf}$ (e.g., ``Security [Agent]'');
  \item A set of \emph{agent tasks} $T = \{t_1, \ldots, t_n\}$ that contribute to both $g_f$ (via means-ends or task decomposition) and $g_{nf}$ (via contribution links).
\end{itemize}
\end{definition}

\begin{definition}[Crosscutting Density]
For a task $t$ in the SR model, let $\mathit{NFR}(t) = \{g_{nf} \mid \exists\, V = (g_f, g_{nf}, T) \text{ with } t \in T\}$ be the set of NFR softgoals that $t$ contributes to. The \emph{crosscutting density} of $t$ is $\delta(t) = |\mathit{NFR}(t)|$.
\begin{itemize}[leftmargin=*]
  \item $\delta(t) = 0$: $t$ has no NFR impact (pure functional task).
  \item $\delta(t) = 1$: $t$ is a single-concern crosscutting point (as in RE 2004).
  \item $\delta(t) \geq 2$: $t$ exhibits \textbf{multi-dimensional crosscutting}, requiring $\delta(t)$ aspects composed at the same join point.
\end{itemize}
\end{definition}

The key distinction from RE 2004 is that agent tasks exhibit high crosscutting density. In the original media shop case study, tasks typically had $\delta \leq 2$. In agent systems, tasks like ``Call LLM Provider'' reach $\delta = 4$ or higher, simultaneously touching security (prompt injection risk), cost management (token consumption), reliability (provider availability), and observability (decision logging). This creates \emph{overlapping V-graphs} that share the same bottom vertex but connect to different NFR softgoals.

High crosscutting density has a practical consequence: when $\delta(t) \geq 3$, the aspects applied at task $t$'s join point must be \emph{composed}, and the composition order matters (e.g., authorization must precede rate limiting, which must precede cost accounting). The pattern composition ordering in Section~\ref{sec:patterns} directly addresses this challenge.

\begin{figure}[t]
\centering
\begin{tikzpicture}[
  goal/.style={rounded rectangle, draw, minimum width=1.4cm, minimum height=0.6cm, font=\scriptsize, align=center},
  softgoal/.style={cloud, cloud puffs=8, cloud puff arc=120, draw, minimum width=1.3cm, minimum height=0.4cm, aspect=2.5, font=\scriptsize, align=center, inner sep=0pt},
  task/.style={rectangle, draw, rounded corners=1pt, minimum width=1.2cm, minimum height=0.5cm, font=\scriptsize, align=center},
  contrib/.style={-{Stealth[length=2mm]}, dashed, font=\tiny},
  decomp/.style={-{Stealth[length=2mm]}},
  >=Stealth
]
\node[goal] (fg) at (-2.5,2.2) {Execute\\Tool};
\node[softgoal] (s1) at (0.5,3.2) {Security};
\node[softgoal] (s2) at (2.5,3.2) {Cost\\Efficiency};
\node[softgoal] (s3) at (0.5,1.2) {Reliability};
\node[softgoal] (s4) at (2.5,1.2) {Observa-\\bility};
\node[task, fill=gray!15, minimum width=1.8cm] (t) at (0,-0.2) {Call LLM\\Provider};
\draw[decomp] (fg) -- (t) node[midway, left, font=\tiny] {means-ends};
\draw[contrib, red] (t) -- (s1) node[midway, left, font=\tiny, red] {$--$};
\draw[contrib, red] (t) -- (s2) node[midway, right, font=\tiny, red] {$--$};
\draw[contrib, red] (t) -- (s3) node[midway, left, font=\tiny, red] {$-$};
\draw[contrib, red] (t) -- (s4) node[midway, right, font=\tiny, red] {$-$};
\draw[dotted, gray] (fg) -- (s1);
\draw[dotted, gray] (fg) -- (s2);
\draw[dotted, gray] (fg) -- (s3);
\draw[dotted, gray] (fg) -- (s4);
\node[font=\tiny\itshape, blue, right] at (3.2,3.2) {PromptGuard};
\node[font=\tiny\itshape, blue, right] at (3.2,2.6) {TokenBudget};
\node[font=\tiny\itshape, blue, right] at (3.2,1.2) {CircuitBreaker};
\node[font=\tiny\itshape, blue, right] at (3.2,0.6) {AuditTrail};
\end{tikzpicture}
\caption{Overlapping V-graphs for agents: the ``Call LLM'' task negatively impacts four NFR softgoals (red dashed arrows), each revealing a crosscutting concern requiring a mitigating aspect.}
\label{fig:vgraph}
\end{figure}
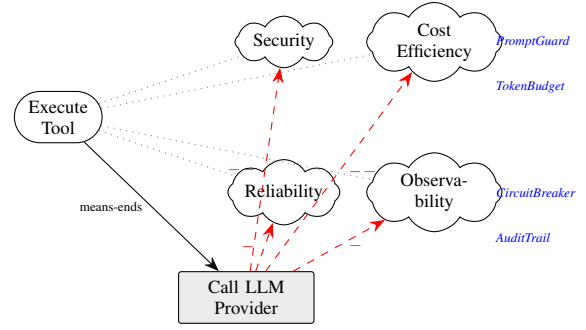

The \emph{AspectFinder for Agents} algorithm extends the original AspectFinder~\cite{yu2004goals} as follows:

\begin{enumerate}[leftmargin=*]
  \item \textbf{Enumerate tasks.} For each agent task $t$ in the SR model, identify all NFR softgoals $\{g_{nf}^1, \ldots, g_{nf}^k\}$ that $t$ contributes to via contribution links.
  \item \textbf{Construct V-graphs.} If $k \geq 1$, each $(g_f, g_{nf}^i, t)$ triple forms a V-graph, revealing a potential aspect. The task's crosscutting density is $\delta(t) = k$. When $\delta(t) \geq 2$, the task produces $k$ overlapping V-graphs, requiring composed aspects at that join point.
  \item \textbf{Group by NFR.} Group V-graphs by NFR softgoal to form \emph{aspect candidates}. Each group collects all tasks that contribute to a given NFR, defining the set of join points for that aspect.
  \item \textbf{Validate via scattering.} For each aspect candidate, measure the degree of concern scattering in the source code (number of modules containing that concern's implementation). A concern scattered across $m \geq 3$ modules confirms the V-graph prediction.
  \item \textbf{Instantiate.} For each validated aspect candidate, determine: the \emph{advice type} (before for precondition checks, after for logging, around for interception), the \emph{join points} (which functions to advise), and the \emph{composition order} relative to other aspects.
\end{enumerate}

Step~4 is our addition to the original algorithm: by grounding the goal-model analysis in empirical source code measurement, we strengthen the validation of discovered aspects and provide quantitative evidence of crosscutting (Table~\ref{tab:scattering}).

\subsection{Phase 3: Pattern Instantiation}
\label{sec:phase3}

Each validated aspect candidate from Phase~2 is instantiated as a pattern in the pattern language. Following the pattern format established by Alexander~\cite{alexander1977pattern} and adapted for software by Gamma et al.~\cite{gamma1995design}, each pattern provides:
\begin{enumerate}[leftmargin=*]
  \item \textbf{Problem}: The crosscutting concern, stated in the agent domain context.
  \item \textbf{\istar{} Goal Model}: The NFR softgoal decomposition showing how the concern is operationalized through tasks and contribution links.
  \item \textbf{Solution}: The aspect implementation in Rust, specifying the advice type (\texttt{before}, \texttt{after}, or \texttt{around}) and the aspect's internal state management.
  \item \textbf{Composition}: Prerequisite and ``composes with'' relationships to other patterns, including the recommended stacking order.
\end{enumerate}

The pattern format deliberately bridges two audiences: (1)~requirements engineers, who work with the \istar{} goal models and NFR categories, and (2)~developers, who work with the Rust aspect implementations. The \istar{} model provides the ``why'' (which NFR is addressed and how it decomposes), while the aspect code provides the ``how'' (concrete, reusable implementation).

\section{Pattern Language}
\label{sec:patterns}

We present 12 patterns organized across four NFR categories (Table~\ref{tab:patterns}). Due to space constraints, we detail four representative patterns---one per category, prioritizing the four novel agent-specific patterns---and summarize the remaining eight.

\begin{table}[t]
\centering
\caption{Pattern language overview: 12 patterns across 4 NFR categories. Patterns marked with $\star$ are novel agent-specific contributions.}
\label{tab:patterns}
\small
\begin{tabular}{@{}llll@{}}
\toprule
\textbf{Category} & \textbf{Pattern} & \textbf{Advice} & \textbf{Source} \\
\midrule
\multirow{4}{*}{Security}
  & Authorization Guard & Before & Existing \\
  & Input Validation & Around & Existing \\
  & Tool Scope Sandbox$^\star$ & Before & Novel \\
  & Prompt Guard$^\star$ & Before & Novel \\
\midrule
\multirow{2}{*}{Reliability}
  & Circuit Breaker & Around & Existing \\
  & Rate Limiter & Around & Existing \\
\midrule
\multirow{4}{*}{Observability}
  & Structured Logger & B/A/AE & Existing \\
  & Performance Monitor & Around & Existing \\
  & Metrics Collector & Around & Existing \\
  & Action Audit Trail$^\star$ & B/A/AE & Novel \\
\midrule
\multirow{2}{*}{Cost Mgmt}
  & Token Budget$^\star$ & Around & Novel \\
  & Response Cache & Around & Existing \\
\bottomrule
\end{tabular}
\end{table}

\subsection{Pattern Format}

Each pattern is presented with: (1)~\textbf{Problem}---the crosscutting concern; (2)~\textbf{\istar{} Goal Model}---the softgoal decomposition; (3)~\textbf{Solution}---the aspect implementation; (4)~\textbf{Composition}---relationships with other patterns.

\subsection{Tool Scope Sandbox (Security, Novel)}
\label{sec:pat-toolscope}

\textbf{Problem.} Autonomous agents with tool-calling capabilities can access files, execute commands, and interact with external services. Without containment, agents may access unauthorized resources---accidentally or through prompt injection attacks.

\textbf{\istar{} Goal Model.} The NFR softgoal \emph{Contain Agent Actions [Agent System]} decomposes into three subgoals: \emph{Enforce Filesystem Boundaries} (tasks: define allowed paths, resolve and check paths), \emph{Filter Executable Commands} (tasks: define command allowlist, parse and match), and \emph{Restrict Network Access} (tasks: define allowed domains, validate URLs). Contribution links: $++$~Security, $++$~Safety, $-$~Capability.

\textbf{Solution.} The \patname{ToolScopeAspect} intercepts tool execution with \texttt{before} advice, checking the requested operation against configured scope boundaries:

\begin{lstlisting}
struct ToolScopeAspect {
  allowed_paths: Vec<PathBuf>,
  command_allowlist: HashSet<String>,
  allowed_domains: Vec<String>,
}
impl Aspect for ToolScopeAspect {
  fn before(&self, ctx: &JoinPoint) {
    // Check tool call against scope
    // Reject if out of bounds
  }
}
#[aspect(ToolScopeAspect::new(config))]
fn execute_shell(cmd: &str) -> Result<Output> {
  Command::new("sh").arg("-c").arg(cmd).output()
}
\end{lstlisting}

\textbf{Composition.} Prerequisite: \patname{AuthorizationGuard} (authenticate before scoping). Composes with: \patname{ActionAuditTrail} (log scope violations), \patname{PromptGuard} (defense in depth).

\subsection{Token Budget Manager (Cost, Novel)}
\label{sec:pat-tokenbudget}

\textbf{Problem.} LLM-based agents consume tokens that incur monetary costs. Agent reasoning loops, tool call chains, and retries can lead to runaway costs without budget enforcement.

\textbf{\istar{} Goal Model.} The softgoal \emph{Cost Efficiency [Agent System]} decomposes into: \emph{Control Token Usage} (tasks: track tokens per call, maintain running total, enforce limits), \emph{Alert on Anomalies} (tasks: warn at threshold, reject above limit), and \emph{Enable Cost Reporting} (tasks: aggregate per session, export for billing). Contribution links: $++$~Sustainability, $++$~Predictability, $-$~Capability.

\textbf{Solution.} The \patname{TokenBudgetAspect} uses \texttt{around} advice to check remaining budget before each LLM call and record usage after:

\begin{lstlisting}
impl Aspect for TokenBudgetAspect {
  fn around(&self, pjp: ProceedingJoinPoint)
    -> Result<Box<dyn Any>, AspectError> {
    let budget = self.budget.lock();
    if budget.used >= budget.max {
      return Err(AspectError::execution(
        "Token budget exceeded"));
    }
    drop(budget);
    let result = pjp.proceed();
    // Record token usage from response
    result
  }
}
\end{lstlisting}

\textbf{Composition.} Prerequisite: \patname{MetricsCollector} (token counting feeds metrics). Composes with: \patname{RateLimiter} (dual cost and rate controls), \patname{ResponseCache} (caching reduces token usage).

\subsection{Action Audit Trail (Observability, Novel)}
\label{sec:pat-audit}

\textbf{Problem.} Autonomous agents make decisions and take actions that affect real systems. For compliance, debugging, and trust, every action must be recorded with full provenance---not just \emph{what} happened, but \emph{why} the agent decided to act.

\textbf{\istar{} Goal Model.} The softgoal \emph{Accountability [Agent System]} decomposes into: \emph{Record All Actions} (tasks: log tool calls with parameters, log LLM decisions), \emph{Preserve Provenance} (tasks: link actions to triggering request, record reasoning chain, timestamp entries), and \emph{Support Compliance Review} (tasks: enable audit queries, detect policy violations). Contribution links: $++$~Trust, $++$~Compliance, $-$~Performance.

\textbf{Solution.} The \patname{AuditTrailAspect} uses \texttt{before}, \texttt{after}, and \texttt{after\_error} advice to record structured audit entries:

\begin{lstlisting}
impl Aspect for AuditTrailAspect {
  fn before(&self, ctx: &JoinPoint) {
    self.store.record(AuditEntry {
      timestamp: Utc::now(),
      session: self.session_id.clone(),
      action: ctx.function_name.into(),
      phase: Phase::Before,
    });
  }
  fn after_error(&self, ctx: &JoinPoint,
                 err: &AspectError) {
    self.store.record(AuditEntry {
      action: ctx.function_name.into(),
      phase: Phase::Error,
      details: format!("{:?}", err),
      ..Default::default()
    });
  }
}
\end{lstlisting}

\textbf{Composition.} Requires: \patname{StructuredLogger} (foundation). Composes with: \patname{AuthorizationGuard} (log who accessed what).

\subsection{Prompt Guard (Security, Novel)}
\label{sec:pat-promptguard}

\textbf{Problem.} LLM-based agents are vulnerable to prompt injection attacks---the \#1 vulnerability in the OWASP Top 10 for LLM Applications~\cite{owasp2025llm}---where malicious inputs manipulate the agent's behavior.

\textbf{\istar{} Goal Model.} The softgoal \emph{Resist Manipulation [Agent System]} decomposes into: \emph{Detect Injection Patterns} (tasks: pattern-match known templates, detect role/system prompt overrides, flag encoding obfuscation), \emph{Respond to Detection} (tasks: log attempt, reject or sanitize, alert operator). Contribution links: $++$~Security, $-$~Usability (false positive risk).

\textbf{Solution.} The \patname{PromptGuardAspect} uses \texttt{before} advice to scan inputs against configurable injection patterns before LLM calls.

\textbf{Composition.} Composes with: \patname{InputValidation} (layered defense), \patname{AuditTrail} (log attempts), \patname{ToolScopeSandbox} (defense in depth).

\subsection{Existing Patterns (Summary)}

The remaining eight patterns adapt well-known crosscutting concerns to the agent domain. While individually familiar from traditional AOP, their agent-specific \istar{} goal models and composition relationships with the four novel patterns constitute part of our contribution.

Table~\ref{tab:patterns} summarizes all eight: \patname{AuthorizationGuard} (RBAC via \texttt{before} advice), \patname{InputValidation} (composable parameter checking via \texttt{around}), \patname{CircuitBreaker} (three-state fault tolerance for LLM providers), \patname{RateLimiter} (token bucket with per-function or global limiting), \patname{StructuredLogger} (entry/exit/error tracking), \patname{PerformanceMonitor} (execution time statistics), \patname{MetricsCollector} (counters and histograms feeding SLA monitoring), and \patname{ResponseCache} (memoization with TTL, reducing token costs). Each adapts a well-known concern to agent-specific join points---e.g., the circuit breaker prevents wasting tokens on a degraded LLM provider, and rate limiting applies to both outbound API calls and inbound gateway requests.

\subsection{Pattern Composition}

Patterns compose through aspect stacking. The recommended composition order for a production agent is:

\begin{lstlisting}
#[aspect(AuthorizationGuard::require_role("op"))]
#[aspect(RateLimitAspect::new(100, 60s))]
#[aspect(LoggingAspect::new())]
#[aspect(AuditTrailAspect::new(store))]
fn call_llm(prompt: &str) -> Result<Response> {
  provider.complete(prompt).await
}
\end{lstlisting}

The full ordering is: authenticate $\rightarrow$ scope $\rightarrow$ guard $\rightarrow$ throttle $\rightarrow$ budget $\rightarrow$ protect $\rightarrow$ observe $\rightarrow$ audit (four annotations shown above for brevity; all eight from Table~\ref{tab:patterns} may be composed). Security-first design dictates that identity verification and scope enforcement precede resource-consuming operations, which precede observability.

\begin{figure}[t]
\centering
\begin{tikzpicture}[
  pat/.style={rectangle, draw, rounded corners=2pt, minimum width=1.4cm, minimum height=0.45cm, font=\tiny, align=center, fill=blue!5},
  novel/.style={pat, fill=yellow!15, thick},
  prereq/.style={-{Stealth[length=2mm]}, thick},
  compose/.style={-{Stealth[length=2mm]}, dashed, gray},
  cat/.style={font=\tiny\bfseries, text=blue!70},
  >=Stealth
]
\node[cat] at (-0.5,3.8) {Security};
\node[pat] (auth) at (0,3.2) {Auth Guard};
\node[novel] (scope) at (0,2.4) {Tool Scope$^\star$};
\node[pat] (val) at (0,1.6) {Validation};
\node[novel] (guard) at (0,0.8) {Prompt Guard$^\star$};

\node[cat] at (2.2,3.8) {Reliability};
\node[pat] (cb) at (2.5,3.2) {Circuit Brkr};
\node[pat] (rl) at (2.5,2.4) {Rate Limiter};

\node[cat] at (4.8,3.8) {Observability};
\node[pat] (log) at (5,3.2) {Logger};
\node[novel] (audit) at (5,2.4) {Audit Trail$^\star$};
\node[pat] (perf) at (5,1.6) {Perf Monitor};
\node[pat] (met) at (5,0.8) {Metrics};

\node[cat] at (7.3,3.8) {Cost};
\node[novel] (tb) at (7.5,3.2) {Token Budget$^\star$};
\node[pat] (cache) at (7.5,2.4) {Resp Cache};

\draw[prereq] (auth) -- (scope);
\draw[prereq] (log) -- (audit);

\draw[compose] (scope.west) to[out=200, in=160] (guard.west);
\draw[compose] (guard.east) -- (audit.west);
\draw[compose] (rl) -- (tb);
\draw[compose] (tb) -- (cache);
\draw[compose] (met.east) to[out=0, in=-90] (tb.south);
\draw[compose] (scope.east) -- (audit.west);
\end{tikzpicture}
\caption{Pattern relationships: solid arrows show prerequisites, dashed arrows show composition relationships. Patterns marked $\star$ are novel.}
\label{fig:patternrel}
\end{figure}
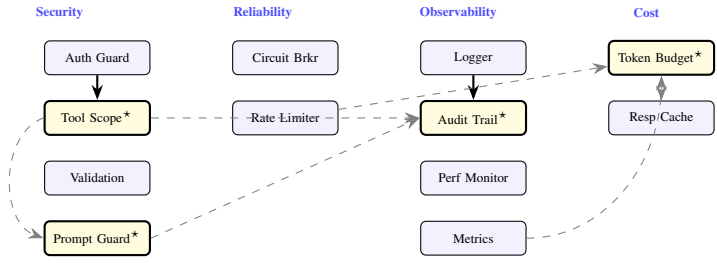

Figure~\ref{fig:patternrel} shows the composition and prerequisite relationships between patterns. Solid arrows indicate prerequisite relationships (the source pattern should execute before the target), and dashed arrows indicate ``composes well with'' relationships.

\section{Case Study: \agentfw{}}
\label{sec:casestudy}

We validate the pattern language by analyzing an open-source autonomous agent framework written in Rust.

\subsection{Subject System}
\label{sec:subject}

\agentfw{}~\cite{zeroclaw2025} is an open-source autonomous AI agent framework written in Rust, comprising 192 source files (129{,}040 LOC) organized into 29 module directories. We selected this framework because it is representative of modern agent architectures while being large enough to exhibit meaningful crosscutting. Its key architectural characteristics include:
\begin{itemize}[leftmargin=*]
  \item \textbf{LLM integration}: Support for 22+ LLM providers (Anthropic, OpenAI, Gemini, Ollama, OpenRouter, and compatible APIs) with reliable routing across providers.
  \item \textbf{Tool capabilities}: 35 tool implementations including shell execution, file read/write, browser control, screenshots, memory operations, web search, hardware memory mapping, and agent delegation.
  \item \textbf{Security}: Gateway pairing with pairing codes, filesystem scoping with allowed paths, command allowlists, authentication subsystem, and rate limiting through a security policy engine.
  \item \textbf{Memory}: Persistent memory subsystem with SQLite storage, markdown processing, embeddings, vector similarity search, RAG, and memory hygiene maintenance.
  \item \textbf{Communication}: 23 channel integrations (Discord, Slack, Telegram, IRC, Matrix, WhatsApp, iMessage, email, DingTalk, Lark, Mattermost, QQ, Signal, ClawdTalk, and others).
  \item \textbf{Infrastructure}: Daemon supervisor with restart backoff, cron scheduler, tunnel providers (ngrok, Cloudflare, Tailscale), health checks, peripheral device support, cost tracking, an HTTP/SSE/WebSocket gateway, and a hook/event dispatch system.
\end{itemize}

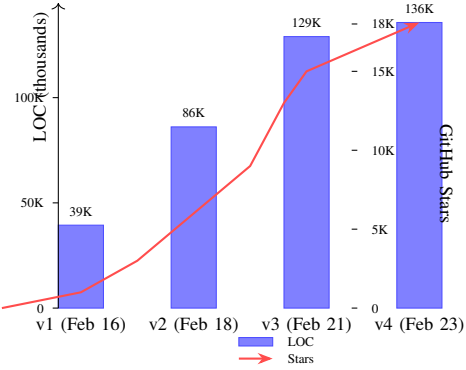
\begin{figure}[t]
\centering
\begin{tikzpicture}[
  xscale=1.5, yscale=0.028,
  growthbar/.style={fill=blue!50, draw=blue!70},
  starline/.style={draw=red!70, thick, -{Stealth[length=2mm]}},
  axis/.style={->, thin}
]
\fill[growthbar] (0,0) rectangle (0.4,39.4);
\node[font=\tiny, above] at (0.2,39.4) {39K};
\fill[growthbar] (1,0) rectangle (1.4,86.1);
\node[font=\tiny, above] at (1.2,86.1) {86K};
\fill[growthbar] (2,0) rectangle (2.4,129.0);
\node[font=\tiny, above] at (2.2,129.0) {129K};
\fill[growthbar] (3,0) rectangle (3.4,135.7);
\node[font=\tiny, above] at (3.2,135.7) {136K};
\node[font=\scriptsize] at (0.2,-8)  {v1 (Feb 16)};
\node[font=\scriptsize] at (1.2,-8)  {v2 (Feb 18)};
\node[font=\scriptsize] at (2.2,-8)  {v3 (Feb 21)};
\node[font=\scriptsize] at (3.2,-8)  {v4 (Feb 23)};
\draw[axis] (0,0) -- (0,145) node[font=\scriptsize, rotate=90, anchor=south, xshift=-28pt] {LOC (thousands)};
\foreach \y/\label in {0/0, 50/50K, 100/100K} {
  \draw (-0.05,\y) -- (0,\y) node[font=\tiny, anchor=east, xshift=-2pt] {\label};
}
\draw[starline] (-0.5, 0) -- (0.2, 7.5) -- (0.7, 22.5) -- (1.2, 45.0) -- (1.7, 67.5) -- (2.0, 97.5) -- (2.2, 112.5) -- (3.2, 135.7);
\node[font=\scriptsize, rotate=-90, anchor=south] at (3.3, 64) {GitHub Stars};
\foreach \y/\label in {0/0, 37.5/5K, 75/10K, 112.5/15K, 135/18K} {
  \draw (2.6,\y) -- (2.65,\y) node[font=\tiny, anchor=west, xshift=2pt] {\label};
}
\fill[growthbar] (1.6,-18) rectangle (1.9,-14);
\node[font=\tiny, anchor=west] at (1.95,-16) {LOC};
\draw[starline] (1.6,-24) -- (1.9,-24);
\node[font=\tiny, anchor=west] at (1.95,-24) {Stars};
\end{tikzpicture}
\caption{ZeroClaw codebase growth (LOC, bars) and community adoption (GitHub stars, line) across four analysis snapshots. The framework grew 3.4$\times$ in seven days while accumulating 18{,}000 GitHub stars, indicating that the crosscutting concerns identified here affect a rapidly growing, widely-adopted system.}
\label{fig:growth}
\end{figure}

This architectural breadth means that crosscutting concerns manifest across diverse module boundaries---from tool handlers to channel integrations to provider backends---providing a rich testbed for aspect analysis.

\subsection{Goal Modeling Results}

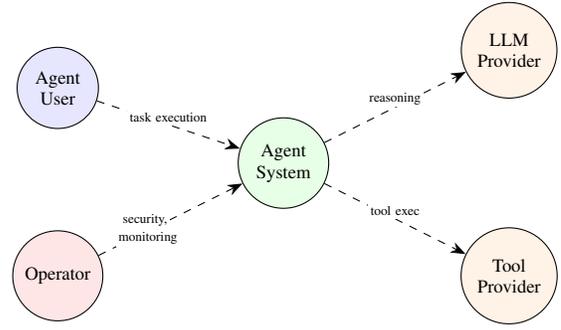
\begin{figure}[t]
\centering
\begin{tikzpicture}[
  actor/.style={circle, draw, minimum size=1cm, font=\scriptsize, align=center},
  dep/.style={-{Stealth[length=2mm]}, dashed},
  dlabel/.style={font=\tiny, fill=white, inner sep=1pt},
  >=Stealth
]
\node[actor, fill=blue!10] (user) at (0,2) {Agent\\User};
\node[actor, fill=green!10] (agent) at (3,1) {Agent\\System};
\node[actor, fill=orange!10] (llm) at (6,2.5) {LLM\\Provider};
\node[actor, fill=orange!10] (tool) at (6,-0.5) {Tool\\Provider};
\node[actor, fill=red!10] (op) at (0,-0.5) {Operator};
\draw[dep] (user) -- (agent) node[dlabel, midway, above] {task execution};
\draw[dep] (agent) -- (llm) node[dlabel, midway, above] {reasoning};
\draw[dep] (agent) -- (tool) node[dlabel, midway, above] {tool exec};
\draw[dep] (op) -- (agent) node[dlabel, midway, left] {security,};
\node[font=\tiny, fill=white, inner sep=1pt] at (1.2,0) {monitoring};
\end{tikzpicture}
\caption{Strategic Dependency model for \agentfw{} with five actors and their goal dependencies.}
\label{fig:sdmodel}
\end{figure}

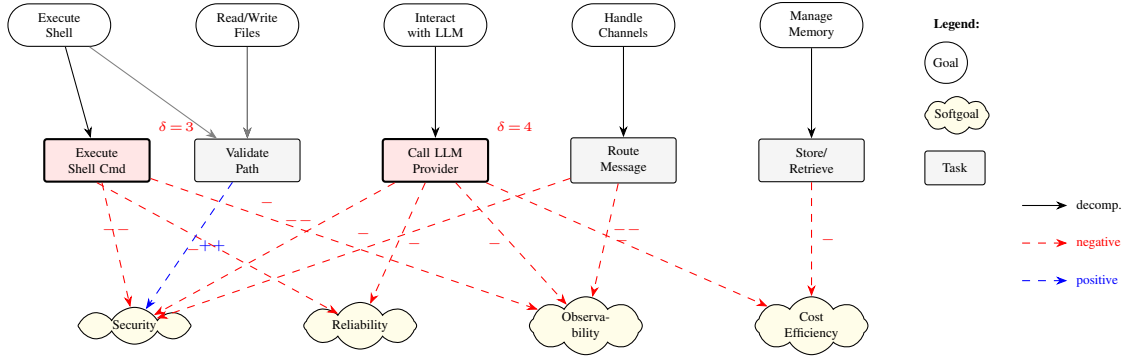
\begin{figure*}[t]
\centering
\begin{tikzpicture}[
  goal/.style={rounded rectangle, draw, minimum width=1.6cm, minimum height=0.55cm, font=\tiny, align=center, fill=white},
  softgoal/.style={cloud, cloud puffs=8, cloud puff arc=120, draw, minimum width=1.5cm, minimum height=0.35cm, aspect=2.5, font=\tiny, align=center, inner sep=0pt, fill=yellow!10},
  task/.style={rectangle, draw, rounded corners=1pt, minimum width=1.4cm, minimum height=0.45cm, font=\tiny, align=center, fill=gray!8},
  hightask/.style={task, fill=red!10, thick},
  actor/.style={rectangle, draw, dashed, rounded corners=5pt, inner sep=8pt},
  decomp/.style={-{Stealth[length=1.5mm]}},
  contrib/.style={-{Stealth[length=1.5mm]}, dashed},
  andlink/.style={decomp},
  >=Stealth,
  every node/.append style={font=\tiny}
]
\node[goal] (g1) at (0,4) {Execute\\Shell};
\node[goal] (g2) at (2.5,4) {Read/Write\\Files};
\node[goal] (g3) at (5,4) {Interact\\with LLM};
\node[goal] (g4) at (7.5,4) {Handle\\Channels};
\node[goal] (g5) at (10,4) {Manage\\Memory};

\node[hightask] (t1) at (0.5,2.2) {Execute\\Shell Cmd};
\node[task] (t2) at (2.5,2.2) {Validate\\Path};
\node[hightask] (t3) at (5,2.2) {Call LLM\\Provider};
\node[task] (t4) at (7.5,2.2) {Route\\Message};
\node[task] (t5) at (10,2.2) {Store/\\Retrieve};

\node[softgoal] (s1) at (1,0) {Security};
\node[softgoal] (s2) at (4,0) {Reliability};
\node[softgoal] (s3) at (7,0) {Observa-\\bility};
\node[softgoal] (s4) at (10,0) {Cost\\Efficiency};

\draw[andlink] (g1) -- (t1);
\draw[andlink] (g2) -- (t2);
\draw[andlink] (g3) -- (t3);
\draw[andlink] (g4) -- (t4);
\draw[andlink] (g5) -- (t5);
\draw[andlink, gray] (g1) -- (t2);
\draw[andlink, gray] (g2) -- (t2);

\draw[contrib, red] (t1) -- (s1) node[midway, above, font=\tiny, red] {$--$};
\draw[contrib, red] (t1) -- (s3) node[pos=0.3, above, font=\tiny, red] {$-$};
\draw[contrib, red] (t1.south) -- (s2) node[pos=0.4, below, font=\tiny, red] {$-$};

\draw[contrib, blue] (t2) -- (s1) node[midway, right, font=\tiny, blue] {$++$};

\draw[contrib, red] (t3) -- (s1) node[pos=0.4, above, font=\tiny, red] {$--$};
\draw[contrib, red] (t3) -- (s2) node[midway, right, font=\tiny, red] {$-$};
\draw[contrib, red] (t3) -- (s3) node[midway, left, font=\tiny, red] {$-$};
\draw[contrib, red] (t3) -- (s4) node[midway, above, font=\tiny, red] {$--$};

\draw[contrib, red] (t4) -- (s1) node[pos=0.5, above, font=\tiny, red] {$-$};
\draw[contrib, red] (t4) -- (s3) node[midway, right, font=\tiny, red] {$-$};

\draw[contrib, red] (t5) -- (s4) node[midway, right, font=\tiny, red] {$-$};

\node[font=\tiny\bfseries, red, above right=-1pt of t1] {$\delta\!=\!3$};
\node[font=\tiny\bfseries, red, above right=-1pt of t3] {$\delta\!=\!4$};

\node[anchor=west, font=\tiny] at (11.5,4) {\textbf{Legend:}};
\node[goal, anchor=west, minimum width=0.8cm] (lg) at (11.5,3.5) {Goal};
\node[softgoal, anchor=west, minimum width=0.8cm] (ls) at (11.5,2.8) {Softgoal};
\node[task, anchor=west, minimum width=0.8cm] (lt) at (11.5,2.1) {Task};
\draw[andlink] (12.8,1.6) -- ++(0.6,0) node[right, font=\tiny] {decomp.};
\draw[contrib, red] (12.8,1.1) -- ++(0.6,0) node[right, font=\tiny] {negative};
\draw[contrib, blue] (12.8,0.6) -- ++(0.6,0) node[right, font=\tiny] {positive};
\end{tikzpicture}
\caption{Partial \istar{} Strategic Rationale model for the Agent System actor, showing 5 functional goals, 5 tasks, and 4 NFR softgoals. Red dashed arrows indicate negative contributions (the task hurts the softgoal); blue dashed arrows indicate positive contributions (the task helps). Tasks with high crosscutting density ($\delta \geq 3$) are highlighted. ``Call LLM Provider'' ($\delta=4$) requires four composed aspects.}
\label{fig:srmodel}
\end{figure*}

We constructed an \istar{} SD model for \agentfw{} with the five actors described in Section~\ref{sec:phase1} (Fig.~\ref{fig:sdmodel}). From this, we constructed a Strategic Rationale (SR) model for the Agent System actor (Fig.~\ref{fig:srmodel}), identifying 8 functional goals, 4 NFR softgoals, and the tasks connecting them. Figure~\ref{fig:srmodel} highlights the two highest-density tasks: \emph{Call LLM Provider} ($\delta=4$) and \emph{Execute Shell} ($\delta=3$).

\subsection{Crosscutting Concern Discovery}

Applying V-graph analysis to the SR model, we identified 28 V-graphs revealing crosscutting concerns across 4 NFR categories. Notably, agent-specific tasks such as ``Call LLM Provider'' and ``Execute Shell Command'' each participate in 4+ V-graphs simultaneously---exhibiting the multi-dimensional crosscutting described in Section~\ref{sec:phase2}.

\textbf{Scattering analysis.} To ground the V-graph analysis in empirical evidence, we measured the degree of crosscutting by counting the number of source files (out of 200 total, spanning 29 module directories across 135{,}651 lines of Rust code) in which each concern's implementation is scattered (Table~\ref{tab:scattering}). Scattering is measured by searching for concern-specific API patterns: security checks (\texttt{allowed}, \texttt{forbidden}, \texttt{deny}, \texttt{pairing}, \texttt{secret}, \texttt{auth}), logging (\texttt{tracing::}, \texttt{info!}, \texttt{warn!}, \texttt{error!}), etc.

\begin{table}[t]
\centering
\caption{Crosscutting concern scattering in \agentfw{} (v4): files and modules containing each concern (out of 200 files across 29 modules).}
\label{tab:scattering}
\small
\begin{tabular}{@{}lccc@{}}
\toprule
\textbf{Concern} & \textbf{Files} & \textbf{Modules} & \textbf{Pattern} \\
\midrule
\multicolumn{4}{@{}l}{\textit{Established concerns}} \\
Error handling & 161 (81\%) & 27 & Circ.\ Breaker \\
Configuration & 128 (64\%) & 27 & --- \\
Security/auth & 104 (52\%) & 23 & Auth.\ Guard \\
Logging & 68 (34\%) & 19 & Struct.\ Logger \\
Path validation & 68 (34\%) & 20 & Tool Scope \\
Retry/resilience & 57 (29\%) & 16 & Circ.\ Breaker \\
Rate limiting & 29 (15\%) & 8 & Rate Limiter \\
\midrule
\multicolumn{4}{@{}l}{\textit{Newly identified in v3, confirmed in v4}} \\
Cost tracking & 42 (21\%) & 14 & Token Budget \\
Hook/dispatch & 37 (19\%) & 13 & --- \\
Telemetry/metrics & 21 (11\%) & 7 & Perf.\ Monitor \\
Approval/HITL & 17 (9\%) & 9 & Action Audit \\
\bottomrule
\end{tabular}
\end{table}

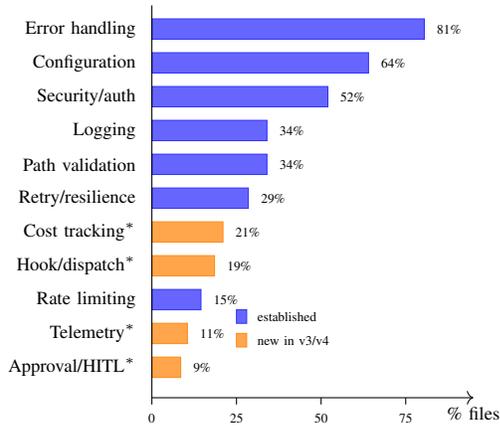
\begin{figure}[t]
\centering
\begin{tikzpicture}[
  xscale=0.045, yscale=0.45,
  established/.style={fill=blue!60, draw=blue!80},
  newbar/.style={fill=orange!70, draw=orange!90},
  label/.style={font=\scriptsize, anchor=east}
]
\node[label] at (-2,10) {Error handling};
\node[label] at (-2,9)  {Configuration};
\node[label] at (-2,8)  {Security/auth};
\node[label] at (-2,7)  {Logging};
\node[label] at (-2,6)  {Path validation};
\node[label] at (-2,5)  {Retry/resilience};
\node[label] at (-2,4)  {Cost tracking$^*$};
\node[label] at (-2,3)  {Hook/dispatch$^*$};
\node[label] at (-2,2)  {Rate limiting};
\node[label] at (-2,1)  {Telemetry$^*$};
\node[label] at (-2,0)  {Approval/HITL$^*$};
\fill[established] (0,9.7) rectangle (80.5,10.3);
\fill[established] (0,8.7) rectangle (64.0,9.3);
\fill[established] (0,7.7) rectangle (52.0,8.3);
\fill[established] (0,6.7) rectangle (34.0,7.3);
\fill[established] (0,5.7) rectangle (34.0,6.3);
\fill[established] (0,4.7) rectangle (28.5,5.3);
\fill[newbar]      (0,3.7) rectangle (21.0,4.3);
\fill[newbar]      (0,2.7) rectangle (18.5,3.3);
\fill[established] (0,1.7) rectangle (14.5,2.3);
\fill[newbar]      (0,0.7) rectangle (10.5,1.3);
\fill[newbar]      (0,-0.3) rectangle (8.5,0.3);
\node[font=\tiny, anchor=west] at (81.5,10) {81\%};
\node[font=\tiny, anchor=west] at (65.0,9)  {64\%};
\node[font=\tiny, anchor=west] at (53.0,8)  {52\%};
\node[font=\tiny, anchor=west] at (35.0,7)  {34\%};
\node[font=\tiny, anchor=west] at (35.0,6)  {34\%};
\node[font=\tiny, anchor=west] at (29.5,5)  {29\%};
\node[font=\tiny, anchor=west] at (22.0,4)  {21\%};
\node[font=\tiny, anchor=west] at (19.5,3)  {19\%};
\node[font=\tiny, anchor=west] at (15.5,2)  {15\%};
\node[font=\tiny, anchor=west] at (11.5,1)  {11\%};
\node[font=\tiny, anchor=west] at (9.5,0)   {9\%};
\draw[->] (0,-0.9) -- (95,-0.9) node[font=\scriptsize, anchor=north] {\% files};
\draw (0,-0.9) -- (0,10.7);
\foreach \x in {0,25,50,75} {
  \draw (\x,-0.9) -- (\x,-1.1) node[font=\tiny, below] {\x};
}
\fill[established] (25,1.3) rectangle (28,1.7);
\node[font=\tiny, anchor=west] at (28.5,1.5) {established};
\fill[newbar] (25,0.6) rectangle (28,1.0);
\node[font=\tiny, anchor=west] at (28.5,0.8) {new in v3/v4};
\end{tikzpicture}
\caption{Crosscutting concern scattering in \agentfw{} (v4, 200 files). Blue: established concerns; orange ($^*$): newly identified in v3, confirmed in v4.}
\label{fig:scattering}
\end{figure}

Five findings are particularly noteworthy. First, \textbf{rate limiting} has three independent implementations: a security policy \texttt{ActionTracker}, a gateway-specific \texttt{SlidingWindowRateLimiter}, and ad-hoc inline checks in tool handlers---each with different error formats and return types. This is a textbook example of a scattered crosscutting concern; the pattern is unchanged across all four versions despite significant codebase growth. Second, \textbf{path validation} to prevent workspace-escape attacks is implemented in three structurally different ways (async vs.\ synchronous canonicalization, delegated vs.\ inline checking, \texttt{ToolResult} vs.\ \texttt{anyhow::bail!} error handling), creating a risk surface where a fix to one site may not propagate to others. Third, the framework defines an \textbf{Observer} trait for structured event reporting (9 files, 2{,}364 LOC), yet 68 files across 19 modules bypass it with direct \texttt{tracing::} calls---demonstrating that even when a modularization exists, the lack of enforcement leads to scattering. Fourth, v3 introduced a dedicated \textbf{hooks} module (5 files) as a native event-dispatch mechanism, making it the closest architectural analogue to AOP join points in the codebase; in v4, 32 of 37 files referencing hook/dispatch patterns still lie outside the module, showing the concern remains mid-migration. Fifth, \textbf{approval/human-in-the-loop} exhibits the worst centralization ratio in the codebase: a dedicated \texttt{approval/} module exists but contains only 1 of 17 files referencing the concept, with 6 scattered across tool handlers, 2 in the security module, and 2 in channel code---unchanged across v3 and v4, confirming this is a structurally persistent gap.

\textbf{Evolution analysis.} We analyzed four versions of \agentfw{} (91~files $\rightarrow$ 161~files $\rightarrow$ 192~files $\rightarrow$ 200~files, 120\% total growth) to examine how scattering evolves as the codebase grows.

\begin{table}[t]
\centering
\caption{Scattering stability across four codebase versions (120\% total file growth). Configuration plateaued after rapid growth; security/auth rose with v4 hardening.}
\label{tab:evolution}
\scriptsize
\begin{tabular}{@{}lrrrrr@{}}
\toprule
\textbf{Concern} & \textbf{v1} & \textbf{v2} & \textbf{v3} & \textbf{v4} & \textbf{$\Delta$} \\
 & \textbf{(91)} & \textbf{(161)} & \textbf{(192)} & \textbf{(200)} & \textbf{v1$\to$v4} \\
\midrule
Error handling    & 79\% & 81\% & 81\% & 81\% & $+$2pp \\
Configuration     & 50\% & 57\% & 64\% & 64\% & $+$14pp \\
Security/auth     & 51\% & 42\% & 50\% & 52\% & $+$1pp \\
Logging           & 37\% & 32\% & 34\% & 34\% & $-$3pp \\
Retry/resilience  & 30\% & 29\% & 29\% & 29\% & $-$1pp \\
Path validation   & 31\% & 29\% & 34\% & 34\% & $+$3pp \\
Rate limiting     & 15\% & 16\% & 15\% & 15\% & $\approx$0pp \\
\bottomrule
\end{tabular}
\end{table}

Table~\ref{tab:evolution} shows that most scattering ratios are remarkably stable across four versions: error handling holds at $\approx$81\%, retry/resilience at $\approx$29\%, and rate limiting at $\approx$15\%, confirming these are structural characteristics of the architecture rather than accidental artifacts of a particular version. \textbf{Configuration} plateaued at 64\% in v3/v4 after growing sharply from 50\% in v1; examination of v3's \texttt{config/traits.rs} reveals a planned decoupling abstraction that has not yet been adopted by consumer modules. \textbf{Security/auth} rose from 50\% to 52\% in v4 as the \texttt{security/} module expanded from 11 to 16 files with new prompt injection defenses---demonstrating that active hardening effort does not reduce scattering but increases absolute file coverage. All new modules in v2--v4 exhibit inherited crosscutting patterns rather than centralizing concerns. This stability implies that crosscutting concern LOC grows \emph{proportionally} with codebase size---aspect-oriented modularization would break this inheritance, as new modules need only aspect annotations rather than copied concern code.

\subsection{Pattern Application}
\label{sec:pattern-app}

We applied the pattern language to \agentfw{} by mapping each discovered crosscutting concern to patterns from our catalog. Table~\ref{tab:coverage} shows the results. All 12 patterns in the catalog are applicable: the four novel agent-specific patterns (\patname{ToolScopeSandbox}, \patname{PromptGuard}, \patname{TokenBudget}, \patname{ActionAuditTrail}) address concerns that currently lack any modular implementation in the framework.

\begin{table}[t]
\centering
\caption{NFR coverage: mapping concerns in \agentfw{} to patterns. Concerns marked with $\dagger$ have no existing modular implementation.}
\label{tab:coverage}
\small
\begin{tabular}{@{}p{1.4cm}p{3.2cm}c@{}}
\toprule
\textbf{NFR} & \textbf{Patterns Applied} & \textbf{Concerns} \\
\midrule
Security & Auth.\ Guard, Validation, Tool Scope$^\star$, Prompt Guard$^\star$ & 2 scattered + 2 absent$^\dagger$ \\
Reliability & Circuit Breaker, Rate Limiter & 2 scattered \\
Observ. & Logger, Perf.\ Monitor, Metrics, Audit Trail$^\star$ & 1 partial + 3 absent$^\dagger$ \\
Cost & Token Budget$^\star$, Resp.\ Cache & 2 absent$^\dagger$ \\
\bottomrule
\end{tabular}
\end{table}

For the scattered concerns (security, logging, rate limiting, retry/resilience, path validation), an aspect-oriented refactoring would consolidate the duplicated implementations into single aspects applied at join points. For the absent concerns (prompt injection defense, token budget management, action audit trails), the pattern language provides ready-to-use aspect designs.

To illustrate the concrete impact, we built a compilable proof-of-concept that faithfully reconstructs \agentfw{}'s scattered rate-limiting code across 8 tool functions, then refactors it using real \aspectrs{} aspects (all tests pass). The current implementation requires developers to manually insert rate-limit checks at each call site with inconsistent APIs:

\begin{lstlisting}
// Current: 3 independent implementations
// tools/shell.rs -- ActionTracker (sliding window)
if tracker.is_rate_limited() {
  return ToolResult { success: false, .. }
}
tracker.record_action();
// gateway/mod.rs -- SlidingWindowRateLimiter
if !limiter.allow_pair(key) { return Err(..); }
// cron/scheduler.rs -- yet another format
if tracker.is_rate_limited() {
  return (false, "rate limit exceeded".into())
}
\end{lstlisting}

With aspect-oriented modularization, all 8 tool functions become:

\begin{lstlisting}
// Refactored: 1 aspect, 8 join points
#[aspect(RateLimitAspect::new(100, 3600s))]
fn execute_shell(cmd: &str) -> Result<..> { .. }
#[aspect(RateLimitAspect::new(100, 3600s))]
fn read_file(path: &str) -> Result<..> { .. }
// ... 4 more tool functions with same pattern
#[aspect(RateLimitAspect::new(10, 60s))]
fn handle_gateway(key: &str) -> Result<..> { .. }
#[aspect(RateLimitAspect::new(30, 3600s))]
fn run_scheduled(name: &str) -> Result<..> { .. }
\end{lstlisting}

Table~\ref{tab:refactoring} quantifies the improvement. The refactoring eliminates approximately 195 lines of scattered, inconsistent rate-limiting code (including \texttt{ActionTracker}, \texttt{SlidingWindowRateLimiter}, \texttt{GatewayRateLimiter} structs and inline checks across 8 functions), replacing them with 8 declarative aspect annotations reusing the existing \patname{RateLimiter} aspect.

\begin{table}[t]
\centering
\caption{Before/after metrics for rate-limiting refactoring (proof-of-concept with 8 tool functions).}
\label{tab:refactoring}
\small
\begin{tabular}{@{}lcc@{}}
\toprule
\textbf{Metric} & \textbf{Before} & \textbf{After} \\
\midrule
Concern-specific LOC & $\sim$195 & 8 \\
Independent implementations & 3 & 1 \\
Algorithms used & 2 & 1 \\
Error format variants & 3 & 1 \\
Functions with inline checks & 8 & 0$^\dagger$ \\
\bottomrule
\multicolumn{3}{l}{\scriptsize $^\dagger$ Concern is encapsulated in the aspect; functions contain only annotations.}
\end{tabular}
\end{table}

Extrapolating from the rate-limiting PoC ($\sim$195 LOC scattered $\rightarrow$ 8 LOC annotations, 96\% reduction), we estimate that aspect-oriented modularization of all seven measured concerns would eliminate approximately 2{,}100--2{,}700 LOC of scattered concern-specific code across the 200-file codebase (extrapolated from the rate-limiting PoC ratio; the v3$\to$v4 growth from 192 to 200 files added proportionally more scattered concern LOC, consistent with the evolution data in Table~\ref{tab:evolution}). Importantly, the evolution data (Table~\ref{tab:evolution}) shows that under continued growth, scattered LOC grows proportionally with codebase size, while aspect-oriented LOC remains near-constant (aspect definitions are $O(1)$; join point annotations are $O(n)$ at $\sim$1 LOC each).

Similarly, path validation---which prevents workspace-escape attacks---is currently implemented in three divergent ways in \agentfw{}. The file-read handler uses async canonicalization and delegates to \texttt{security.is\_resolved\_path\_allowed()}; the file-write handler canonicalizes the \emph{parent} directory after creation; and the skills module uses synchronous \texttt{canonicalize()} with direct \texttt{starts\_with()} checking and \texttt{anyhow::bail!} for errors. The \patname{ToolScopeSandbox} pattern would replace all three with a uniform \texttt{before} advice that applies the same validation strategy at every tool execution join point, eliminating the risk surface created by inconsistent implementations.

\subsection{Discussion of Results}

The scattering analysis confirms that \agentfw{}, despite being a well-structured Rust codebase organized into 29 modules, exhibits significant crosscutting: every concern we measured spans at least 8 modules, with error handling touching 27 modules and configuration access appearing in 27. The overlap between concerns---many files contain checks for multiple concerns---further illustrates the tangling problem.

Table~\ref{tab:vgraphs} maps each V-graph from the goal model to the empirically measured scattering in the source code. The table reveals two categories of results: concerns with measured scattering (confirming the V-graph identification) and concerns that are entirely \emph{absent} from the implementation despite being identified by V-graph analysis.

\begin{table}[t]
\centering
\caption{V-graph analysis results for \agentfw{}. Each row maps a V-graph $(g_f, g_{nf}, T)$ to empirically observed scattering. ``Absent'' indicates a concern identified by V-graph analysis that has no implementation in the framework.}
\label{tab:vgraphs}
\scriptsize
\begin{tabular}{@{}p{1.6cm}p{1.2cm}p{2.4cm}p{2.2cm}@{}}
\toprule
\textbf{Func.\ Goal} & \textbf{NFR} & \textbf{Task(s)} & \textbf{Scattering} \\
\midrule
Exec Tool & Security & Check auth, validate path & 104 files (52\%) \\
Exec Tool & Safety & Scope filesystem & 68 files (34\%) \\
Call LLM & Reliability & Retry on failure & 57 files (29\%) \\
Serve Req. & Perf. & Rate limit & 29 files (15\%) \\
Handle Ch. & Observ. & Log events & 68 files (34\%) \\
Exec Tool & Cost & Track tokens & 42 files (21\%) \\
\midrule
Exec Tool & Observ. & Audit actions & Absent$^\dagger$ \\
Call LLM & Security & Guard prompt & Absent$^\dagger$ \\
\bottomrule
\multicolumn{4}{l}{\scriptsize $^\dagger$ Concern identified by V-graph analysis but absent from implementation.}
\end{tabular}
\end{table}

The results demonstrate that V-graph analysis identifies both \emph{implemented but scattered} concerns and \emph{missing} concerns. Six of eight V-graphs correspond to crosscutting concerns that are implemented but scattered---confirming that these are genuine crosscutting concerns amenable to aspect modularization. Notably, the V-graph connecting \emph{Execute Tool} to \emph{Security} through \emph{Check Authorization} and \emph{Validate Path} correctly identifies scattering across 104 files in 23 modules in v4---a growth reflecting active security hardening. Significantly, the \emph{Cost} V-graph---which was absent in v2---corresponds to 42 scattered files in v4, confirming that the framework has begun addressing token-budget tracking but without modular enforcement. Two V-graphs still identify concerns with \emph{no} implementation, revealing persistent requirements gaps.

The four novel patterns address a gap: \agentfw{} currently has \emph{no modular} implementation of prompt injection defense or structured action audit trails; token budget tracking has emerged (42 files, 21\% in v4) but remains scattered rather than aspect-encapsulated. These are precisely the agent-specific crosscutting concerns that the V-graph extension (overlapping V-graphs) uniquely identifies through multi-dimensional crosscutting analysis.

\textbf{Comparison with RE 2004.} The original media shop case study~\cite{yu2004goals} identified three crosscutting aspects (security, usability, portability) from a goal model with 4 actors and approximately 20 goals. Our agent case study identifies 11 crosscutting concerns (Table~\ref{tab:scattering}) from a goal model with 5 actors, 8 functional goals, and 7 softgoals, with empirical validation across 200 source files and four codebase versions. The agent domain exhibits more pervasive crosscutting: the most scattered concern (error handling, 81\%) affects nearly four times more modules than any concern in the media shop case study. Furthermore, the four-version longitudinal analysis provides evidence that crosscutting concern density is \emph{structurally stable}---the scattering ratios do not diminish as the codebase grows, suggesting that without architectural intervention, they compound.

\section{Discussion}
\label{sec:discussion}

\subsection{Benefits of Aspect-Oriented Agentic AI}

The pattern language offers several benefits for agent development:

\textbf{Modularity.} Each NFR is encapsulated in a single aspect, eliminating scattering and tangling. In \agentfw{}, the rate limiting concern currently spans 29 files across 8 modules with three independent implementations; the \patname{RateLimiter} pattern consolidates this into a single reusable aspect applied declaratively at each join point.

\textbf{Consistency.} Aspects enforce uniform behavior across all join points. The path validation concern in \agentfw{} uses three structurally different implementations (async vs.\ sync, delegated vs.\ inline, \texttt{ToolResult} vs.\ \texttt{anyhow::bail!}). The \patname{ToolScopeSandbox} aspect would enforce a single, consistent validation strategy.

\textbf{Evolvability.} New crosscutting concerns can be added as new aspects without modifying the agent core. When a new NFR emerges (e.g., GDPR data handling), a new aspect can be developed and applied to existing join points without changing agent logic.

\textbf{Evolutionary resilience.} Our four-version evolution analysis (Table~\ref{tab:evolution}) provides empirical evidence that scattering ratios remain stable as the codebase grows: new modules inherit the same ad-hoc patterns as existing ones. Under aspect-oriented modularization, this inheritance is broken: new tool implementations need only add \texttt{\#[aspect(...)]} annotations rather than copying scattered rate-limiting, security, or logging code. The aspect definition itself is maintained in one location; a bug fix or policy change propagates to all join points automatically. The rate-limiting PoC demonstrates this concretely: adding a new tool to \agentfw{} currently requires copying 12--15 LOC of rate-limit checks; with \aspectrs{}, it requires one annotation.

\textbf{Composability.} Aspects stack declaratively via annotations, with a principled ordering (authenticate $\rightarrow$ scope $\rightarrow$ guard $\rightarrow$ throttle $\rightarrow$ budget $\rightarrow$ protect $\rightarrow$ observe $\rightarrow$ audit). The composition order is explicit in the source code, unlike the implicit ordering of ad-hoc inline checks.

\textbf{Traceability.} The three-phase methodology creates a traceable chain: \istar{} softgoals $\rightarrow$ V-graph analysis $\rightarrow$ aspect patterns $\rightarrow$ Rust implementations. This supports requirements traceability from early-phase goals to deployed code.

\subsection{Revisiting RE 2004 After 22 Years}

The V-graph model remains applicable in the agentic AI domain, but agent tasks exhibit significantly higher \emph{multi-dimensional crosscutting} than the original media shop case study (Section~\ref{sec:phase2}). Where ``Process Order'' might crosscut 2--3 softgoals, ``Call LLM Provider'' simultaneously crosscuts 5+ (security, cost, reliability, observability, safety), creating overlapping V-graphs that make systematic aspect discovery critical---ad-hoc identification risks missing concerns entirely, as evidenced by the absence of prompt injection defense and token budget management in \agentfw{}.

Beyond aspect \emph{discovery}, the pattern language provides a catalog of reusable \emph{aspect implementations} that practitioners can directly apply, bridging the gap between requirements analysis and software design.

\subsection{Implications for Practitioners}

The pattern language has immediate practical implications for agent system development:

\textbf{Proactive NFR identification.} Rather than discovering crosscutting concerns after they become maintenance problems (as in \agentfw{}, where rate limiting was independently implemented three times), the goals-to-aspects methodology enables proactive identification during requirements analysis. The V-graph analysis systematically reveals \emph{every} crosscutting concern in the goal model, preventing the common situation where concerns like prompt injection defense or token budget management are overlooked entirely.

\textbf{Architecture decision support.} The pattern language provides a decision framework for agent architects. Given an agent's \istar{} goal model, the pattern catalog maps each NFR softgoal to a concrete aspect implementation. The composition ordering (Fig.~\ref{fig:patternrel}) provides guidance on how to layer multiple aspects without conflicting interactions.

\textbf{Regulatory compliance.} As AI governance regulations emerge (EU AI Act~\cite{euaiact2024}, NIST AI RMF), the \patname{ActionAuditTrail} and \patname{ToolScopeSandbox} patterns provide concrete traceability and containment mechanisms, creating auditable evidence that NFRs have been systematically addressed.

\textbf{Sustainability.} The \patname{TokenBudget} and \patname{ResponseCache} patterns support computational sustainability by controlling token consumption and reducing unnecessary LLM API calls---particularly important given that unconstrained agent reasoning loops can consume thousands of tokens per turn.

\subsection{Threats to Validity}

\textbf{Internal validity.} The mapping from goal models to patterns involves expert judgment; different analysts might identify different V-graphs and thus different aspects. We mitigate this threat in two ways. First, we ground each pattern in empirical source code analysis: every concern we identify through V-graph analysis is validated by measuring its scattering across modules (Table~\ref{tab:scattering}). Second, the four novel patterns address concerns explicitly identified in the OWASP LLM Top 10~\cite{owasp2025llm} and agent architecture surveys~\cite{wang2024survey}, providing independent validation of their relevance.

\textbf{External validity.} We analyze a single agent framework. While \agentfw{} is representative---it supports 22 LLM providers, multiple tool types, and 24 communication channels, spanning 200 files and 135{,}651 LOC---generalization to other frameworks (particularly those in Python and TypeScript) requires further case studies. However, the crosscutting concerns we identify (security, observability, cost management, resilience) are architectural in nature and not language-specific, suggesting the pattern language is broadly applicable.

\textbf{Construct validity.} Our scattering metric (file count per concern) is based on keyword-pattern matching, which introduces both false positives and false negatives. To assess measurement precision, we manually validated a sample of three concerns. For \emph{security/authorization} (25 of 104 files sampled, including 5 files from new modules), manual inspection yielded 80\% precision---false positives arose from external API credential handling (e.g., sending keys to third-party APIs) matching the \texttt{auth} pattern. For \emph{rate limiting} (14 of 29 files inspected), precision was 64\%---false positives included comments \emph{about} external rate limits and temporal scheduling described as ``throttling.'' For \emph{path validation} (19 of 68 files sampled), precision was only 32\%---the \texttt{starts\_with} pattern is heavily overloaded in Rust (URL scheme detection, token format checking, protocol parsing). Applying these precision rates yields adjusted estimates of approximately 83, 19, and 22 files respectively. The qualitative conclusions---that these concerns are scattered across multiple modules and exhibit inconsistent implementations---hold under the adjusted numbers, but the raw counts in Table~\ref{tab:scattering} should be interpreted as upper bounds. Finer-grained metrics (e.g., co-occurrence of \texttt{canonicalize} and \texttt{starts\_with} within the same function for path validation) would improve precision in future work. As additional construct validation, we built a compilable proof-of-concept that faithfully reconstructs and then refactors the rate-limiting concern using real \aspectrs{} aspects, confirming the feasibility of the proposed modularization (Table~\ref{tab:refactoring}).

\section{Related Work}
\label{sec:related}

\textbf{Goal-Oriented RE.} The \istar{} framework~\cite{yu1997modelling,dalpiaz2016istar} and the NFR framework~\cite{chung2000nfr} provide the theoretical foundation for our work. The NFR framework's treatment of softgoals and contribution links is central to our V-graph analysis. Ernst et al.~\cite{ernst2012agile} extended goal modeling to agile and paraconsistent reasoning contexts, demonstrating the adaptability of goal-oriented approaches to new development paradigms. Our work similarly adapts goal modeling to the agentic AI paradigm.

\textbf{Aspect-Oriented RE.} Rashid et al.~\cite{rashid2003modularization} established AORE as a discipline, proposing methods for modularizing and composing aspectual requirements. Their comprehensive treatment~\cite{rashid2013aore} provides frameworks for identifying, specifying, and managing crosscutting concerns at the requirements level. Baniassad and Clarke~\cite{baniassad2004discovering} proposed theme-based methods for discovering early aspects from requirements documents. Moreira et al.~\cite{moreira2005aore} explored multi-dimensional separation of concerns in RE. Our work extends these foundations to the agentic AI domain with four novel agent-specific patterns and empirical validation through source code analysis.

\textbf{Goals to Aspects.} Yu et al.~\cite{yu2004goals} proposed the goals-to-aspects methodology that we extend, demonstrating V-graph analysis on a media shop case study. Niu et al.~\cite{niu2009aspects} extended this methodology across the full software lifecycle, from requirements to design to implementation. Sampaio et al.~\cite{sampaio2005eaminer} developed EA-Miner, a tool for automating aspect-oriented requirements identification using NLP techniques. Our contribution differs in three ways: (1)~we apply the methodology to the novel domain of agentic AI, (2)~we extend the V-graph model to capture multi-dimensional crosscutting, and (3)~we bridge the gap from aspect discovery to aspect implementation through a pattern language.

\textbf{RE for AI/ML Systems.} Belani et al.~\cite{belani2019requirements} identified unique RE challenges for AI-based complex systems, including data-dependent quality attributes and evolving requirements. Habibullah and Horkoff~\cite{habibullah2021requirements} studied how practitioners handle NFRs for ML systems, finding that fairness, explainability, and safety are poorly addressed. Ahmad et al.~\cite{ahmad2023requirements} provided a systematic mapping study covering 126 primary studies on RE for AI systems. Our work complements these by providing a \emph{concrete} pattern language for modularizing agent NFRs through aspect-oriented techniques, rather than only identifying challenges.

\textbf{LLM Agent Architectures.} Wang et al.~\cite{wang2024survey} and Xi et al.~\cite{xi2024agentgym} surveyed LLM-based autonomous agents, identifying planning, memory, and tool use as key capabilities. Yao et al.~\cite{yao2023react} proposed the ReAct framework for synergizing reasoning and acting. Shinn et al.~\cite{shinn2023reflexion} introduced Reflexion for verbal reinforcement learning. The OWASP LLM Top 10~\cite{owasp2025llm} identifies critical security risks including prompt injection (\#1), insecure output handling (\#2), and excessive agency (\#8). Our \patname{PromptGuard}, \patname{ToolScopeSandbox}, and \patname{ActionAuditTrail} patterns directly address these top risks through aspect-oriented modularization, providing a systematic RE-grounded approach to agent security.

\section{Conclusion}
\label{sec:conclusion}

We have revisited the goals-to-aspects methodology proposed at RE 2004 and extended it to the agentic AI domain. Our three contributions address a timely gap. First, the pattern language of 12 reusable patterns---including four novel agent-specific patterns for tool-scope sandboxing, prompt injection detection, token budget management, and action audit trails---provides a principled, immediately actionable approach for identifying and modularizing crosscutting concerns in agent systems. Second, the V-graph extension to overlapping V-graphs captures the multi-dimensional crosscutting that characterizes agent tasks, where a single task like ``Call LLM Provider'' simultaneously crosscuts security, cost, reliability, and observability. Third, the case study on \agentfw{} (200~files, 135{,}651~LOC, 29~modules) provides quantitative evidence of concern scattering across four codebase versions, with every measured concern spanning 8--27 modules.

The results suggest that aspect-oriented techniques are not merely applicable but \emph{essential} for the agentic AI domain, where the density of crosscutting concerns exceeds that of traditional software. The pattern language bridges the gap between early-phase requirements analysis and concrete implementation, offering practitioners both ``what aspects are needed'' and ``how to implement them.''

Future work includes: (1)~multi-agent extensions addressing inter-agent crosscutting concerns (e.g., shared token budgets, coordinated audit trails); (2)~automated pattern recommendation from goal models, extending EA-Miner~\cite{sampaio2005eaminer} to the agent domain; (3)~multiple case studies across diverse agent frameworks (Python, TypeScript, Rust) to validate pattern generalizability; and (4)~empirical studies measuring the impact of aspect-oriented modularization on agent system maintainability and defect rates.

\subsection*{Data Availability Statement}
The replication package (pattern language, goal models, scattering measurement data, and aspect implementations) is available on Zenodo at \url{https://doi.org/10.5281/zenodo.18724794}. The subject agent framework \agentfw{} is open-source at \url{https://github.com/zeroclaw-labs/zeroclaw}; the four analysis snapshots are pinned to commits \texttt{21dc22f} (v1, 2026-02-16), \texttt{d42cb1e} (v2, 2026-02-18), \texttt{38e27ff} (v3, 2026-02-21), and \texttt{aa45c30} (v4, 2026-02-23). Scattering measurements are fully reproducible using the \texttt{aspect-analysis.sh} script included in the replication package. The AOP framework \aspectrs{} is available at \url{https://crates.io/crates/aspect-core}.

\subsection*{AI Tool Disclosure}
In accordance with IEEE policy, we disclose that AI coding assistants were used to support LaTeX formatting. All research design, goal modeling, pattern identification, and analysis were performed by the authors.

\bibliographystyle{IEEEtran}
\bibliography{references}

\end{document}